# Being Bayesian about Network Structure

**Nir Friedman**
School of Computer Science & Engineering
Hebrew University
Jerusalem, 91904, Israel
*nir@cs.huji.ac.il*

**Daphne Koller**
Computer Science Department
Stanford University
Stanford, CA 94305-9010
*koller@cs.stanford.edu*

## Abstract

In many domains, we are interested in analyzing the structure of the underlying distribution, e.g., whether one variable is a direct parent of the other. Bayesian model-selection attempts to find the MAP model and use its structure to answer these questions. However, when the amount of available data is modest, there might be many models that have non-negligible posterior. Thus, we want compute the Bayesian posterior of a feature, i.e., the total posterior probability of all models that contain it. In this paper, we propose a new approach for this task. We first show how to efficiently compute a sum over the exponential number of networks that are consistent with a fixed ordering over network variables. This allows us to compute, for a given ordering, both the marginal probability of the data and the posterior of a feature. We then use this result as the basis for an algorithm that approximates the Bayesian posterior of a feature. Our approach uses an *Markov Chain Monte Carlo* (MCMC) method, but over orderings rather than over network structures. The space of orderings is much smaller and more regular than the space of structures, and has a smoother posterior "landscape". We present empirical results on synthetic and real-life datasets that compare our approach to full model averaging (when possible), to MCMC over network structures, and to a non-Bayesian bootstrap approach.

## 1 Introduction

In the last decade there has been a great deal of research focused on the problem of learning Bayesian networks (BNs) from data [3, 8]. An obvious motivation for this problem is to learn a model that we can then use for inference or decision making, as a substitute for a model constructed by a human expert. In certain cases, however, our goal is to learn a model of the system not for prediction, but for discovering the domain structure. For example, we might want to use Bayesian network learning to understand the mechanisms by which genes in a cell produce proteins, which in turn cause other genes to express themselves, or prevent them from doing so [6]. In this case, our main goal is to discover the causal and dependence relations between the expression levels of different genes [12].

The common approach to discovering structure is to use learning with model selection to provide us with a single high-scoring model. We then use that model (or its equivalence class) as our model for the structure of the domain. Indeed, in small domains with a substantial amount of data, it has been shown that the highest scoring model is orders of magnitude more likely than any other [11]. In such cases, model selection is a good approximation. Unfortunately, there are many domains of interest where this situation does not hold. In our gene expression example, it is now possible to measure of the expression levels of thousands of genes in one experiment [12] (where each gene is a random variable in our model [6]), but we typically have only a few hundred of experiments (each of which is a single data case). In cases, like this, where the amount of data is small relative to the size of the model, there are likely to be many models that explain the data reasonably well. Model selection makes an arbitrary choice between these models, and therefore we cannot be confident that the model is a true representation of the underlying process.

Given that there are many qualitatively different structures that are approximately equally good, we cannot learn a unique structure from the data. However, there might be certain features of the domain, e.g., the presence of certain edges, that we can extract reliably. As an extreme example, if two variables are very strongly correlated (e.g., deterministically related to each other), it is likely that an edge between them will appear in any high-scoring model. Our goal, therefore, is to compute how likely a feature such as an edge is to be present over all models, rather than a single model selected by the learning algorithm. In other words, we are interested in computing:

$$P(f \mid D) = \sum_G P(G \mid D) f(G), \qquad (1)$$

where $f(G)$ is 1 if the feature holds in $G$ and 0 otherwise.

The number of BN structures is super-exponential in the number of random variables in the domain; therefore, this summation can be computed in closed form only for very small domains, or those in which we have additional constraints that restrict the space (as in [11]). Alternatively, this summation can be approximated by considering only a subset of possible structures. Several approximations have been proposed [13, 14]. One theoretically well-founded approach is to use Markov Chain Monte Carlo (MCMC) methods: we define a Markov chain over structures whose stationary distribution is the posterior $P(G \mid D)$, we then generate samples from this chain, and use them to estimate Eq. (1).

In this paper, we propose a new approach for evaluating the Bayesian posterior probability of certain structural network properties. Our approach is based on two main ideas. The first is an efficient closed form equation for summing over all networks with at most $k$ parents per node (for some constant $k$) that are consistent with a fixed ordering over the nodes. This equation allows us both to compute the overall probability of the data for this set of networks, and to compute the posterior probability of certain structural features — edges and Markov blankets— over this set. The second idea is the use of an MCMC approach, but over orderings of the network variables rather than directly on BN structures.

The space of orderings is much smaller than the space of network structures; it also appears to be much less peaked, allowing much faster *mixing* (i.e., convergence to the stationary distribution of the Markov chain). We present empirical results illustrating this observation, showing that our approach has substantial advantages over direct MCMC over BN structures. The Markov chain over orderings mixes much faster and more reliably than the chain over network structures. Indeed, different runs of MCMC over networks typically lead to very different estimates in the posterior probabilities of structural features, illustrating poor convergence to the stationary distribution; by contrast, different runs of MCMC over orderings converge reliably to the same estimates. We also present results showing that our approach accurately captures dominant features even with sparse data, and that it outperforms both MCMC over structures and the non-Bayesian bootstrap of [5].

## 2 Bayesian learning of Bayesian networks

Consider the problem of analyzing the distribution over some set of random variables $X_1, \ldots, X_n$, based on a fully observed data set $D = \{\mathbf{x}[1], \ldots, \mathbf{x}[M]\}$, where each $\mathbf{x}[j]$ is a complete assignment to the variables $X_1, \ldots, X_n$.

### 2.1 The Bayesian learning framework

The Bayesian learning paradigm tells us that we must define a prior probability distribution $P(\mathcal{B})$ over the space of possible Bayesian networks $\mathcal{B}$. This prior is then updated using Bayesian conditioning to give a posterior distribution $P(\mathcal{B} \mid D)$ over this space.

For Bayesian networks, the description of a model $\mathcal{B}$ has two components: the structure $G$ of the network, and the values of the numerical parameters $\boldsymbol{\theta}_G$ associated with it. For example, in a discrete Bayesian network of structure $G$, the parameters $\boldsymbol{\theta}_G$ define a multinomial distribution $\boldsymbol{\theta}_{X_i \mid \mathbf{u}}$ for each variable $X_i$ and each assignment of values $\mathbf{u}$ to $\mathrm{Pa}_G(X_i)$. If we consider Gaussian Bayesian networks over continuous domains, then $\boldsymbol{\theta}_{X_i \mid \mathbf{u}}$ contains the coefficients for a linear combination of $\mathbf{u}$ and a variance parameter.

To define the prior $P(\mathcal{B})$, we need to define a discrete probability distribution over graph structures $G$, and for each possible graph $G$, to define a continuous distribution over the set of parameters $\boldsymbol{\theta}_G$.

The prior over structures is usually considered the less important of the two components. Unlike other parts of the posterior, it does not grow as the number of data cases grows. Hence, relatively little attention has been paid to the choice of structure prior, and a simple prior is often chosen largely for pragmatic reasons. The simplest and therefore most common choice is a uniform prior over structures [8]. An alternative prior, and the one we use in our experiments, considers the number of options in determining the families of $G$. If we decide that a node $X_i$ has $k$ parents, then there are $\binom{n-1}{k}$ possible parents sets. If we assume that we choose uniformly from these, we get a prior:

$$P(G) \propto \prod_{i=1}^{n} \binom{n-1}{|\mathrm{Pa}_G(X_i)|}^{-1}.$$

Note that the negative logarithm of this prior corresponds to the *description length* of specifying the parent sets, assuming that the cardinality of these sets are known. Thus, we implicitly assume that cardinalities of parent sets are uniformly distributed.

A key property of all these priors is that they satisfy:

- **Structure modularity** The prior $P(G)$ can be written in the form

$$P(G) \propto \prod_i \rho(X_i, \mathrm{Pa}_G(X_i)).$$

That is, the prior decomposes into a product, with a term for each family in $G$. In other words the choices of the families for the different nodes are independent a priori.

Next we consider the prior over parameters, $P(\boldsymbol{\theta}_G \mid G)$. Here, the form of the prior varies depending on the type of parametric families we consider. In discrete networks, the standard assumption is a *Dirichlet* prior over $\boldsymbol{\theta}_{X_i \mid \mathbf{u}}$ for each node $X_i$ and each instantiation $\mathbf{u}$ to its parents [8]. In Gaussian networks, we might use a Wishart prior [9]. For our purpose, we need only require that the prior satisfies two basic assumptions, as presented in [10]:

- **Global parameter independence:** Let $\boldsymbol{\theta}_{X_i \mid \mathrm{Pa}_G(X_i)}$ be the parameters specifying the behavior of the variable $X_i$ given the various instantiations to its parents. Then we require that

$$P(\boldsymbol{\theta}_G \mid G) = \prod_i P(\boldsymbol{\theta}_{X_i \mid \mathrm{Pa}_G(X_i)} \mid G) \qquad (2)$$

- **Parameter modularity:** Let $G$ and $G'$ be two graphs in which $\mathrm{Pa}_G(X_i) = \mathrm{Pa}_{G'}(X_i) = \mathbf{U}$ then

$$P(\boldsymbol{\theta}_{X_i \mid \mathbf{U}} \mid G) = P(\boldsymbol{\theta}_{X_i \mid \mathbf{U}} \mid G') \qquad (3)$$

Once we define the prior, we can examine the form of the posterior probability. Using Bayes rule, we have that

$$P(G \mid D) \propto P(D \mid G) P(G).$$

The term $P(D \mid G)$ is the *marginal likelihood* of the data given $G$, and is defined the integration over all possible parameter values for $G$.

$$P(D \mid G) = \int P(D \mid G, \boldsymbol{\theta}_G) P(\boldsymbol{\theta}_G \mid G) d\boldsymbol{\theta}_G$$

The term $P(D \mid G, \boldsymbol{\theta}_G)$ is simply the probability of the data given a specific Bayesian network. When the data is *complete*, this is simply a product of conditional probabilities.

Using the above assumptions, one can show (see [10]):

**Theorem 2.1:** *If* $D$ *is complete and* $P(G)$ *satisfies* parameter independence *and* parameter modularity *, then*

$$P(D \mid G) = \prod_i \text{score}(X_i, \text{Pa}_G(X_i) \mid D).$$

*where* $\text{score}(X_i, \mathbf{U} \mid D)$ *is*

$$\int \prod_m P(x_i[m] \mid \mathbf{u}[m], \boldsymbol{\theta}_{X_i|\mathbf{U}}) P(\boldsymbol{\theta}_{X_i|\mathbf{U}}) d\boldsymbol{\theta}_{X_i|\mathbf{U}}$$

If the prior also satisfies structure modularity, we can also conclude that posterior probability decomposes:

$$P(G \mid D) \propto \prod_i \rho(X_i, \text{Pa}_G(X_i)) \mathit{score}(X_i, \text{Pa}_G(X_i) \mid D).$$

### 2.2 Bayesian model averaging

Recall that our goal is to compute the posterior probability of some feature $f(G)$ over all possible graphs $G$. This is equal to:

$$P(f \mid D) = \sum_G f(G) P(G \mid D)$$

The problem, of course, is that the number of possible BN structures is super-exponential: $2^{O(n^2 \log n)}$, where $n$ is the number of variables.

We can reduce this number by restricting attention to structures $G$ where there is a bound $k$ on the number of parents per node. This assumption, which we will make throughout this paper, is a fairly innocuous one. There are few applications in which very large families are called for, and there is rarely enough data to support robust parameter estimation for such families. From a more formal perspective, networks with very large families tend to have low score. Let $\mathcal{G}_k$ be the set of all graphs with indegree bounded by $k$. Note that the number of structures in $\mathcal{G}_k$ is still super-exponential: at least $2^{kn \log n}$.

Thus, exhaustive enumeration over the set of possible BN structures is feasible only for tiny domains (4–5 nodes). One solution, proposed by several researchers [11, 13, 14], is to approximate this exhaustive enumeration by finding a set $\mathcal{G}$ of high scoring structures, and then estimating the relative mass of the structures in $\mathcal{G}$ that contains $f$:

$$P(f \mid D) \approx \frac{\sum_{G \in \mathcal{G}} P(G \mid D) f(G)}{\sum_{G \in \mathcal{G}} \Pr(G \mid D)}. \qquad (4)$$

This approach leaves open the question of how we construct $\mathcal{G}$. The simplest approach is to use model selection to pick a single high-scoring structure, and then use that as our approximation. If the amount of data is large relative to the size of the model, then the posterior will be sharply peaked around a single model, and this approximation is a reasonable one. However, as we discussed in the introduction, there are many interesting domains (e.g., our biological application) where the amount of data is small relative to the size of the model. In this case, there is usually a large number of high-scoring models, so using a single model as our set $\mathcal{G}$ is a very poor approximation.

A simple approach to finding a larger set is to record all the structures examined during the search, and return the high scoring ones. However, the set of structures found in this manner is quite sensitive to the search procedure we use. For example, if we use greedy hill-climbing, then the set of structures we will collect will all be quite similar. Such a restricted set of candidates also show up when we consider multiple restarts of greedy hill-climbing and beam-search. This is a serious problem since we run the risk of getting estimates of confidence that are based on a biased sample of structures.

Madigan and Raftery [13] propose an alternative approach called *Occam's window*, which rejects models whose posterior probability is very low, as well as complex models whose posterior probability is not substantially better than a simpler model (one that contains a subset of the edges). These two principles prune the space of models considered, often to a number small enough to be exhaustively enumerated. Madigan and Raftery also provide a search procedure for finding these models.

An alternative approach, proposed by Madigan and York [14], is based on the use of *Markov chain Monte Carlo (MCMC)* simulation. In this case, we define a Markov Chain over the space of possible structures, whose stationary distribution is the posterior distribution $P(G \mid D)$. We then generate a set of possible structures by doing a random walk in this Markov chain. Assuming that we continue this process until the chain mixes, we can hope to get a set of structures that is representative of the posterior. However, it is not clear how rapidly this type of chain mixes for large domains. The space of structures is very large, and the probability distribution is often quite peaked, with neighboring structures having very different scores. Hence, the mixing rate of the Markov chain can be quite slow.

## 3 Closed form for known ordering

In this section, we temporarily turn our attention to a somewhat easier problem. Rather than perform model averaging over the space of *all* structures, we restrict attention to structures that are consistent with some known total ordering $\prec$. In other words, we restrict attention to structures $G$ where if $X_i \in \text{Pa}_G(X_j)$ then $i \prec j$. This assumption was a standard one in the early work on learning Bayesian networks from data [4].

### 3.1 Computing marginal likelihood

We first consider the problem of computing the probability of the data given the ordering:

$$P(D \mid \prec) = \sum_{G \in \mathcal{G}_k} P(G \mid \prec) P(D \mid G) \tag{5}$$

Note that this summation, although restricted to networks with bounded indegree and consistent with $\prec$, is still exponentially large: the number of such structures is still at least $2^{kn \log n}$.

The key insight is that, when we restrict attention to structures consistent with a given ordering $\prec$, the choice of family for one node places no additional constraints on the choice of family for another. Note that this property does not hold without the restriction on the ordering; for example, if we pick $X_i$ to be a parent of $X_j$, then $X_j$ cannot in turn be a parent of $X_i$.

Therefore, we can choose a structure $G$ consistent with $\prec$ by choosing, independently, a family $\mathbf{U}$ for each node $X_i$. The parameter modularity assumption Eq. (3) states that the choice of parameters for the family of $X_i$ is independent of the choice of family for another family in the network. Hence, summing over possible graphs consistent with $\prec$ is equivalent to summing over possible choices of family for each node, each with its parameter prior. Given our constraint on the size of the family, the possible parent sets for the node $X_i$ is

$$\mathcal{U}_{i,\prec} = \{\mathbf{U} : \mathbf{U} \prec X_i, |\mathbf{U}| \leq k\}.$$

where $\mathbf{U} \prec X_i$ is defined to hold when all nodes in $\mathbf{U}$ precede $X_i$ in $\prec$. Given that, we have

$$\begin{aligned} & P(D \mid \prec) \\ & \propto \sum_{G \in \mathcal{G}_k} \prod_i \rho(X_i, \mathrm{Pa}_G(X_i)) \prod_i score(X_i, \mathrm{Pa}_G(X_i) \mid D) \\ & = \prod_i \sum_{\mathbf{U} \in \mathcal{U}_{i,\prec}} \rho(X_i, \mathbf{U}) score(X_i, \mathbf{U} \mid D) . \end{aligned} \tag{6}$$

Intuitively, the equality states that we can sum over all networks consistent with $\prec$ by summing over the set of possible families for each node, and then multiplying the results for the different nodes. This transformation allows us to compute $P(D \mid \prec)$ very efficiently. The expression on the right-hand side consists of a product with a term for each node $X_i$, each of which is a summation over all possible families for $X_i$. Given the bound $k$ over the number of parents, the number of possible families for a node $X_i$ is at most $\binom{n}{k} \leq n^k$. Hence, the total cost of computing Eq. (6) is at most $n \cdot n^k = n^{k+1}$.

We note that the decomposition of Eq. (6) was first mentioned by Buntine [2], but the ramifications for Bayesian model averaging were not pursued. The concept of Bayesian model averaging using a closed-form summation over an exponentially large set of structures was proposed (in a different setting) in [17].

The computation of $P(D \mid \prec)$ is useful in and of itself; as we show in the next section, computing the probability $P(D \mid \prec)$ is a key step in our MCMC algorithm.

### 3.2 Probabilities of features

For certain types of features $f$, we can use the same technique to compute, in closed form, the probability $P(f \mid \prec, D)$ that $f$ holds in a structure given the ordering and the data.

In general, if $f(\cdot)$ is a feature. We want to compute

$$P(f \mid \prec, D) = \frac{P(f, D \mid \prec)}{P(D \mid \prec)}.$$

We have just shown how to compute the denominator. The numerator is a sum over all structures that contain the feature and are consistent with the ordering:

$$P(f, D \mid \prec) = \sum_{G \in \mathcal{G}_k} f(G) P(G \mid \prec) P(D \mid G) \tag{7}$$

The computation of this term depends on the specific type of feature $f$.

The simplest situation is when we want to compute the posterior probability of the $f_{X_i \to X_j}$ that denotes an edge $X_i \to X_j$. In this case, we can apply the same closed form analysis to (7). The only difference is that we restrict $\mathcal{U}_{j,\prec}$ to consist only of subsets that contain $X_i$. Since the terms that sum over the parents of $X_k$ for $k \neq j$ are not disturbed by this constraint, they cancel out from the equation.

**Proposition 3.1:** $P(f_{X_i \to X_j} \mid \prec, D) =$

$$\frac{\sum_{\{\mathbf{U} \in \mathcal{U}_{i,\prec} : \mathbf{U} \ni X_j\}} \rho(X_i, \mathbf{U}) \mathrm{score}(X_i, \mathbf{U} \mid D)}{\sum_{\mathbf{U} \in \mathcal{U}_{i,\prec}} \rho(X_i, \mathbf{U}) \mathrm{score}(X_i, \mathbf{U} \mid D)}$$

The same argument can be extended to ask more complex queries about the parents of $X_i$. For example, we can compute the posterior probability of a particular choice of parents, as

$$\begin{aligned} & P(\mathrm{Pa}_G(X_i) = \mathbf{U} \mid D, \prec) = \\ & \qquad \frac{\rho(X_i, \mathbf{U}) score(X_i, \mathbf{U} \mid D)}{\sum_{\mathbf{U}' \in \mathcal{U}_{i,\prec}} \rho(X_i, \mathbf{U}') score(X_i, \mathbf{U}' \mid D)}. \end{aligned} \tag{8}$$

A somewhat more subtle computation is required to compute the posterior of $f_{X_i \overset{M}{\sim} X_j}$, the feature that denotes that $X_i$ is in the Markov blanket of $X_j$. Recall this is the case if $G$ contains the edge $X_i \to X_j$, or the edge $X_j \to X_i$, or there is a variable $X_k$ such that both edges $X_i \to X_k$ and $X_j \to X_k$ are in $G$.

Assume, without loss of generality, that $X_i$ precedes $X_j$ in the ordering. In this case, $X_i$ can be in $X_j$'s Markov blanket either if there is an edge from $X_i$ to $X_j$, or if $X_i$ and $X_j$ are both parents of some third node $X_l$. We have just shown how the first of these probabilities $p_j = P(f_{X_i \to X_j} \mid D, \prec)$, can be computed in closed form. We can also easily compute the probability $q_l = P(X_i, X_j \in \mathrm{Pa}_G(X_l) \mid D, \prec)$ that both $X_i$ and $X_j$ are parents of $X_l$: we simply restrict $\mathcal{U}_{l,\prec}$ to families that contain both $X_i$ and $X_j$. The key is to note that as the choice of families of different nodes are independent, these are all independent events. Hence, $X_i$ and $X_j$ are not in the same Markov blanket only if all of these events fail to occur. Thus,

**Proposition 3.2:**

$$P(f_{X_i \overset{M}{\sim} X_j} \mid D, \prec) = 1 - (1 - p_j) \cdot \prod_{X_l \succ X_j} (1 - q_l)$$

Unfortunately, this approach cannot be used to compute the probability of arbitrary structural features. For example, we cannot compute the probability that there exists some directed path from $X_i$ to $X_j$, as we would have to consider all possible ways in which this path could manifest through our exponentially many structures.

We can overcome this difficulty using a simple sampling approach. Eq. (8) provides us with a closed form expression for the exact posterior probability of the different possible families of the node $X_i$. We can therefore easily sample entire networks from the posterior distribution given the ordering: we simply sample a family for each node, according to the distribution in Eq. (8). We can then use the sampled networks to evaluate any feature, such as the existence of a causal path from $X_i$ to $X_j$.

## 4 MCMC methods

In the previous section, we made the simplifying assumption that we were given a predetermined ordering. Although this assumption might be reasonable in certain cases, it is clearly too restrictive in domains where we have very little prior knowledge (e.g., our biology domain). We therefore want to consider structures consistent with all $n!$ possible orderings over BN nodes. Here, unfortunately, we have no elegant tricks that allow a closed form solution. Therefore, we provide a solution which uses our closed form solution of Eq. (6) as a subroutine in a Markov Chain Monte Carlo algorithm [15].

### 4.1 The basic algorithm

We introduce a uniform prior over orderings $\prec$, and define $P(G \mid \prec)$ to be of the same nature as the priors we used in the previous section. It is important to note that the resulting prior over structures has a different form than our original prior over structures. For example, if we define $P(G \mid \prec)$ to be uniform, we have that $P(G)$ is not uniform: graphs that are consistent with more orderings are more likely; for example, a Naive Bayes graph is consistent with $(n-1)!$ orderings, whereas any chain-structured graph is consistent with only one. While this discrepancy in priors is unfortunate, it is important to see it in proportion. The standard priors over network structures are often used not because they are particularly appropriate for a task, but rather because they are simple and easy to work with. In fact, the ubiquitous uniform prior over structures is far from uniform over PDAGs (Markov equivalence classes over network structures) — PDAGs consistent with more structures have a higher induced prior probability. One can argue that, for causal discovery, a uniform prior over PDAGs is more appropriate; nevertheless, a uniform prior over networks is most often used for practical reasons. Finally, the prior induced over our networks does have some justification: one can argue that a structure which is consistent with more orderings makes fewer assumptions about causal ordering, and is therefore more likely a priori.

We now construct a Markov chain $\mathcal{M}$ with state space $\mathcal{O}$, consisting of all $n!$ orderings $\prec$; our construction will guarantee that $\mathcal{M}$ has the stationary distribution $P(\prec \mid D)$. We can then simulate this Markov chain, obtaining a sequence of samples $\prec_1, \ldots, \prec_T$. We can now approximate the expected value of any function $g(\prec)$ as:

$$\boldsymbol{E}[g \mid D] \approx \frac{1}{T} \sum_{t=1}^{T} g(\prec_t).$$

Specifically, we can let $g(\prec)$ be $P(f \mid \prec, D)$ for some feature (edge) $f$. We can then compute $g(\prec_t) = P(f \mid \prec_t, D)$, as described in the previous section.

It remains only to discuss the construction of the Markov chain. We use a standard Metropolis algorithm [15]. We need to guarantee two things:

- that the chain is reversible, i.e., the probability $P(\prec \to \prec') = P(\prec' \to \prec)$;
- that the stationary probability of the chain is the desired posterior $P(\prec \mid D)$.

We accomplish this goal using a standard Metropolis sampling. For each ordering $\prec$, we define a *proposal probability* $q(\prec' \mid \prec)$, which defines the probability that the algorithm will "propose" a move from $\prec$ to $\prec'$. The algorithm then *accepts* this move with probability

$$\min\left[1, \frac{P(\prec' \mid D) q(\prec \mid \prec')}{P(\prec \mid D) q(\prec' \mid \prec)}\right].$$

It is well known that the resulting chain is reversible and has the desired stationary distribution [7].

We consider several specific constructions for the proposal distribution, based on different neighborhoods in the space of orderings. In one very simple construction, we consider only operators that flip two nodes in the ordering (leaving all others unchanged):

$$(i_1 \ldots i_j \ldots i_k \ldots i_n) \mapsto (i_1 \ldots i_k \ldots i_j \ldots i_n).$$

Another operator is "cutting the deck" in the ordering:

$$(i_1 \ldots i_j i_{j+1} \ldots i_n) \mapsto (i_{j+1} \ldots i_n i_1 \ldots i_j).$$

We note that these two types of operators are qualitatively very different. The "flip" operator takes much smaller steps in the space, and is therefore likely to mix much more slowly. However, any single step is substantially more efficient to compute (see below). In our implementation, we choose a flip operator with some probability $p$, and a cut operator with probability $1 - p$. We then pick each of the possible instantiations uniformly (i.e., given that we have decided to cut, all $n$ positions are equally likely).

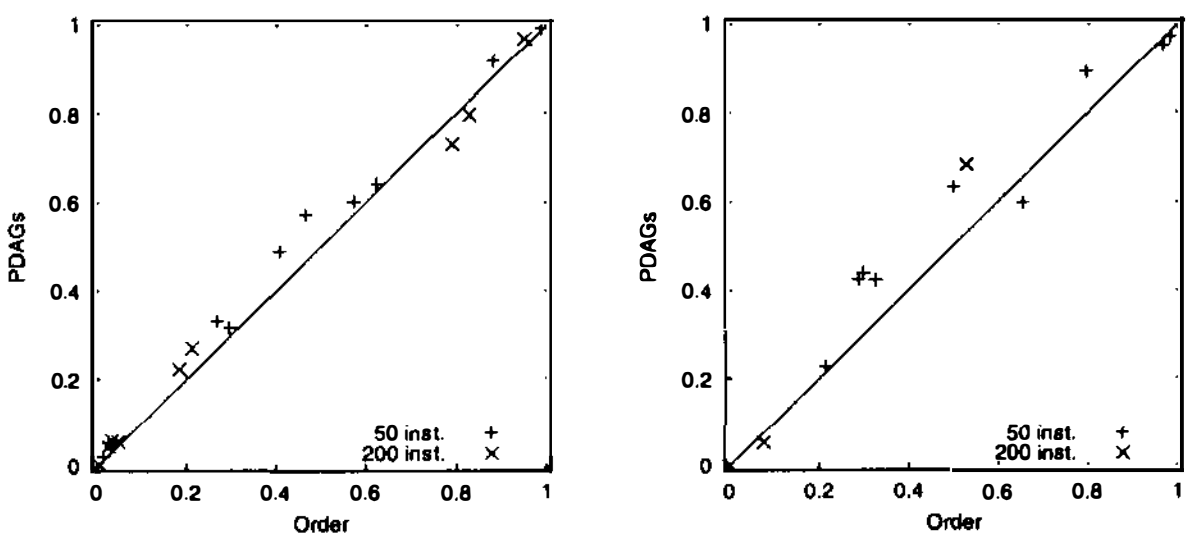

*Flare* *Vote*

Figure 1: Comparison of posterior probabilities for different Markov features between full Bayesian averaging using: orderings ($x$-axis) versus PDAGs ($y$-axis) for two UCI datasets (5 variables each).

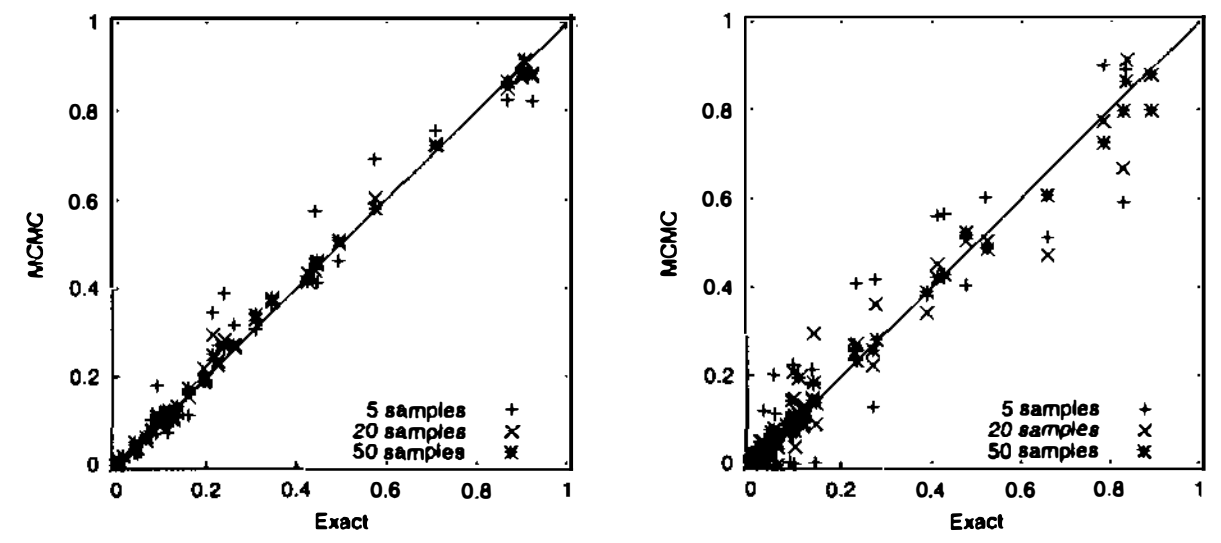

Markov Edges

Figure 2: Comparison of posterior probabilities using true posterior over orderings ($x$-axis) versus ordering-MCMC ($y$-axis). The figures show Markov features and Edge features in the Flare dataset with 100 samples.

### 4.2 Computational tricks

Although the computation of the marginal likelihood is polynomial in $n$, it can still be quite expensive, especially for large networks and reasonable size $k$. We utilize several computational tricks for reducing the complexity of this computation.

First, for each node $X_i$, we restrict attention to at most $m_P$ other nodes as possible parents (for some fixed $m_P$). We select these $m_P$ nodes in advance, before any MCMC step, as follows: for each potential parent $X_j$, we compute the score of the single edge $X_j \rightarrow X_i$; we then select the $m_P$ nodes $X_j$ for which this score was highest.

Second, for each node $X_i$, we precompute the score for some number $m_F$ of the highest-scoring families. Again, this procedure is executed once, at the very beginning of the process. The list of highest-scoring families is sorted in decreasing order; let $\ell_i$ be the score of the worst family in $X_i$'s list. As we consider a particular ordering, we extract from the list all families consistent with that ordering. We know that all families not in the list score no better than $\ell_i$. Thus, if the best family extracted from the list is some factor $\gamma$ better than $\ell_i$, we choose to restrict attention to the families extracted from the list, under the assumption that other families will have negligible effect relative to these high-scoring families. If the best family extracted is not that good, we do a full enumeration.

Third, we prune the exhaustive enumeration of families by ignoring families that augment low-scoring families with low-scoring edges. Specifically, assume that for some family $\mathbf{U}$, we have that $score(X_i, \mathbf{U} \mid D)$ is substantially lower than other families enumerated so far. In this case, families that extend $\mathbf{U}$ are likely to be even worse. More precisely, we define the *incremental value* of a parent $Y$ for $X_i$ to be its added value as a single parent: $\Delta(Y; X_i) = score(X_i, Y) - score(X_i)$. If we now have a family $\mathbf{U}$ such that, for all other possible parents $Y$, $score(X_i, \mathbf{U}) + \Delta(Y; X_i)$ is lower than the best family found so far for $X_i$, we prune all extensions of $\mathbf{U}$.

Finally, we note that when we take a single MCMC step in the space, we can often preserve much of our computation. In particular, let $\prec$ be an ordering and let $\prec'$ be the ordering obtained by flipping $i_j$ and $i_k$. Now, consider the terms in Eq. (6); those terms corresponding to nodes $i_\ell$ in the ordering $\prec$ that precede $i_j$ or succeed $i_k$ do not change, as the set of potential parent sets $\mathcal{U}_{i_l,\prec}$ is the same. Furthermore, the terms for $i_l$ that are between $i_j$ and $i_k$ also have a lot in common — all parent sets $\mathbf{U}$ that contain neither $i_j$ nor $i_k$ remain the same. Thus, we only need to subtract

$$\sum_{\{\mathbf{U} \in \mathcal{U}_{i,\prec} \,:\, \mathbf{U} \ni X_{i_j}\}} \rho(X_i, \mathbf{U}) score(X_i, \mathbf{U} \mid D)$$

and add

$$\sum_{\{\mathbf{U} \in \mathcal{U}_{i,\prec'} \,:\, \mathbf{U} \ni X_{i_k}\}} \rho(X_i, \mathbf{U}) score(X_i, \mathbf{U} \mid D).$$

By contrast, the "cut" operator requires that we recompute the entire summation over families for each variable $X_i$.

## 5 Experimental Results

We first compared the exact posterior computed by summing over *all* orderings to the posterior computed by summing over all equivalence classes of Bayesian networks (PDAGs). (I.e., we counted only a single representative network for each equivalence class.) The purpose of this evaluation is to try and evaluate the effect of the somewhat different prior over structures. Of course, in order to do the exact Bayesian computation we need to do an exhaustive enumeration of hypotheses. For orderings, this enumeration is possible for as many as 10 variables, but for structures, we are limited to domains with 5–6 variables. We took two data sets — *Vote* and *Flare* — from the UCI repository [16] and selected five variables from each. We generated datasets of sizes 50 and 200, and computed the full Bayesian averaging posterior for these datasets using both methods. Figure 1 compares the results for both datasets. We see that for small amounts of data, the two approaches are slightly different but in general quite well correlated. This illustrates that, at least for small data sets, the effect of our different prior does not dominate the posterior value.

Next, we compared the estimates made by our MCMC sampling over orderings to estimates given by the full

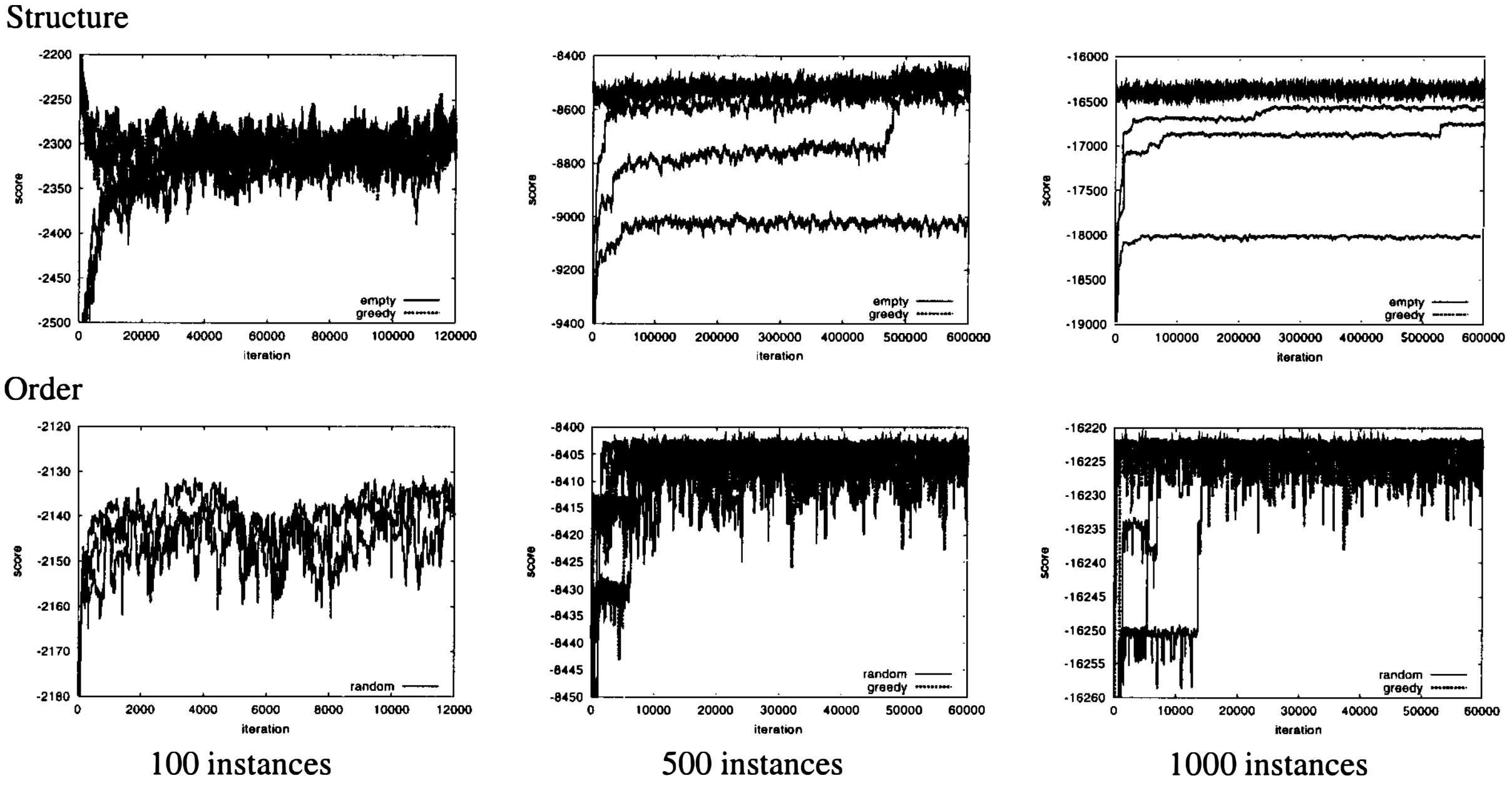

Figure 3: Plots of the progression of the MCMC runs. Each graph shows plots of 6 independent runs over *Alarm* with either 100, 500, and 1000 samples. The graph plot the score ($\log_2(P(D \mid G)P(G))$ or $\log_2(P(D \mid \prec)P(\prec))$) of the "current" candidate ($y$-axis) for different iterations ($x$-axis) of the MCMC sampler. In each plot, three of the runs are seeded with the network found by greedy hill climbing search over network structures. The other three runs are seed either by the empty network in the case of the structure-MCMC or a random ordering in the case of ordering-MCMC.

Bayesian averaging over networks. We experimented on the nine-variable "flare" dataset. We ran the MCMC sampler with a burn-in period of 1,000 steps and then sampled every 100 steps; we experimented with collecting 5, 20, and 50 samples. (We note that these parameters are probably excessive, but they ensure that we are sampling very close the stationary probability of the process.) The results are shown in Figure 2. As we can see, the estimates are very robust. In fact, for Markov features even a sample of 5 orderings gives a surprisingly decent estimate. This is due to the fact that a single sample of an ordering contains information about exponentially many possible structures. For edges we obviously need more samples, as edges that are not in the direction of the ordering necessarily have probability 0. With 20 and 50 samples we see a very close correlation between the MCMC estimate and the exact computation for both types of features.

We then considered larger datasets, where exhaustive enumeration is not an option. For this purpose we used synthetic data generated from the *Alarm* BN [1], a network with 37 nodes. Here, our computational tricks are necessary. We used the following settings: $k$ (max. number of parents in a family) $= 3$; $m_P$ (max. number of potential parents) $= 20$; $m_F$ (number of families cached) $= 4000$; and $\gamma$ (difference in score required in pruning) $= 10$. Note that $\gamma = 10$ corresponds to a difference of $2^{10}$ in the posterior probability of the families. We note that different families have huge differences in score, so a difference of $2^{10}$ in the posterior probability is not uncommon.

Here, our primary goal was the comparison of structure-MCMC and ordering-MCMC. For the structure MCMC, we used a burn in of 100,000 iterations and then sampled every 25,000 iterations. For the order MCMC, we used a burn in of 10,000 iterations and then sampled every 2,500 iterations. In both methods we collected a total of 50 samples per run. One phenomenon that was quite clear was that ordering-MCMC runs *mixed* much faster. That is, after a small number of iterations, these runs reached a "plateau" where successive samples had comparable scores. Runs started in different places (including random ordering and orderings seeded from the results of a greedy-search model selection) rapidly reached the same plateau. On the other hand, MCMC runs over network structures reached very different levels of scores, even though they were run for much larger number of iterations. Figure 3 illustrates this phenomenon for examples of *alarm* with 100, 500, and 1000 instances. Note the substantial difference in scale between the two sets of graphs.

In the case of 100 instances, both MCMC samplers seemed to mix. The structure based sampler mixes after about 20,000–30,000 iterations while the ordering based sampler mixes after about 1,000–2,000 iterations. On the other hand, when we examine 500 samples, the ordering-MCMC converges to a high-scoring plateau, which we believe is the stationary distribution, within 10,000 iterations. By contrast, different runs of the structure-MCMC stayed in very different regions of the in the first 500,000 iterations. The situation is even worse in the case of 1,000 in-stances. In this case the structure based MCMC sampler that starts from an empty network does not reach the level of score achieved by the runs starting from the structure found by greedy hill climbing search. Moreover, these latter runs seem to fluctuate around the score of the initial seed. Note that runs show differences of 100 – 500 bits. Thus, the sub-optimal runs sample from networks that are at least $2^{100}$ less probable!

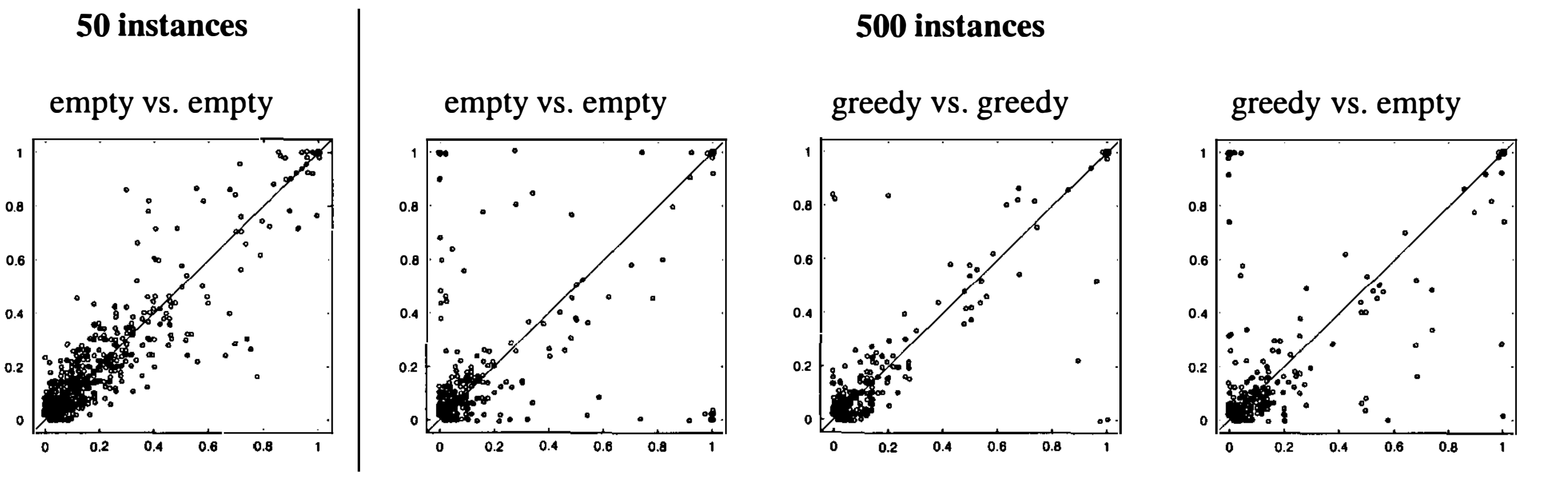

Figure 4: Scatter plots that compare posterior probability of Markov features on the *Alarm* dataset, as determined by different runs of structure-MCMC. Each point corresponds to a single Markov feature; its $x$ and $y$ coordinates denote the posterior estimated by the two compared runs. The position of points is slightly randomly perturbed to visualize clusters of points in the same position.

This phenomenon has two explanations. Either the seed structure is the global optimum and the sampler is sampling from the posterior distribution, which is "centered" around the optimum; or the sampler is stuck in a local "hill" in the space of structures from which it cannot escape. This latter hypothesis is supported by the fact that runs starting at other structures (e.g., the empty network) take a very long time to reach similar level of scores, indicating that there is a very different part of the space on which stationary behavior is reached.

We can provide further support for this second hypothesis by examining the posterior computed for different features in different runs. Figure 4 compares the posterior probability of Markov features assigned by different runs of structure-MCMC. Although different runs give a similar probability estimate to most structural features, there are several features on which they differ radically. In particular, there are features that are assigned probability close to 1 by samples from one run and probability close to 0 by samples from the other. While this behavior is less common in the runs seeded with the greedy structure, it occurs even there. This suggests that each of these runs (even runs that start at the same place) gets trapped in a different local neighborhood in the structure space.

By contrast, comparison of the predictions of different runs of the order based MCMC sampler are tightly correlated. Figure 5 compares two runs, one starting from an ordering consistent with the greedy structure and the other from a random order. We can see that the predictions are very similar, both for the small dataset and the larger one. This observation reaffirms our claim that these different runs are indeed sampling from similar distributions. That is, they are sampling from the true posterior.

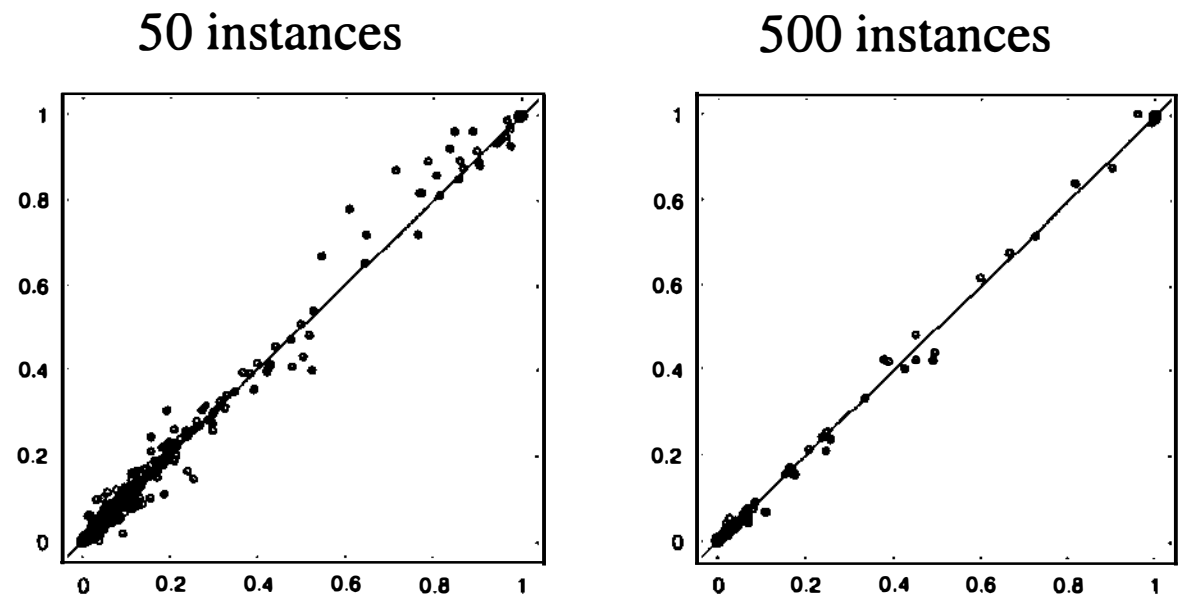

Figure 5: Scatter plots that compare posterior probability of Markov features on the *Alarm* domain as determined by different runs of ordering-MCMC. Each point corresponds to a single Markov feature; its $x$ and $y$ coordinates denote the posterior estimated by the greedy seeded run and a random seeded run respectively.

We believe that the difference in mixing rate is due to the smoother posterior landscape of the space of orderings. In the space of networks, even a small perturbation to a network can lead to a huge difference in score. By contrast, the score of an ordering is a lot less sensitive to slight perturbations. For one, the score of each ordering is an aggregate of the scores of a very large space of structures; hence, differences in scores of individual networks can often cancel out. Furthermore, for most orderings, we are likely to find a consistent structure which is not too bad a fit to the data; hence, an ordering is unlikely to be uniformly horrible.

The disparity in mixing rates is more pronounced for larger datasets. The reason is quite clear: as the amount of data grows, the posterior landscape becomes "sharper" since the effect of a single change on the score is amplified across many samples. As we discussed above, if our dataset is large enough, model selection is often a good approximation to model averaging. (Although this is not quite the case for 1000-instance *Alarm*.) Conversely, if we consider *Alarm* with only 100 samples, or the (fairly small) genetics data set, graphs such as Figure 3 indicate that structure-MCMC does eventually converge (although still more slowly than ordering-MCMC).

We note that, computationally, structure-MCMC is faster than ordering-MCMC. In our current implementation, generating a successor network is about an order of magnitude faster than generating a successor ordering. We therefore designed the runs in Figure 3 to take roughly the same amount of computation time. Thus, even for the same amount of computation, ordering-MCMC mixes faster.

When both ordering-MCMC and structure-MCMC mix, it is possible to compare their estimates. In Figure 6 we see such comparisons for *Alarm*. We see that, in general, the estimates of the two methods are not too far apart, although the posterior estimate of the structure-MCMC is usually larger. This difference between the two approaches raises the obvious question: which estimate is better? Clearly, we cannot compute the exact posterior for a domain of this size, so we cannot answer this question exactly. However, we can test whether the posteriors computed by the different methods can reconstruct features of the generating model. To do so, we label Markov features in the *Alarm* domain as *positive* if they appear in the generating network and *negative* if they do not. We then use our posterior to try and distinguish "true" features from "false" ones: we pick a threshold $t$, and predict that the feature $f$ is "true" if $P(f) > t$. Clearly, as we vary the the value of $t$, we will get different sets of features. At each threshold value we can have two types of errors: *false positives* — positive features that are misclassified as negative, and *false negatives* — negative features that are classified as positive. Different values of $t$ achieve different tradeoffs between these two type of errors. Thus, for each method we can plot the *tradeoff curve* between the two types of errors. Note that, in most applications of structure discovery, we care more about false positives than about false negatives. For example, in our biological application, false negatives are only to be expected — it is unrealistic to expect that we would detect all causal connections based on our limited data. However, false positives correspond to hypothesizing important biological connections spuriously. Thus, our main concern is with the left-hand-side of the tradeoff curve, the part where we have a small number of false positives. Within that region, we want to achieve the smallest possible number of false negatives.

We computed such tradeoff curves for *Alarm* data set with 100 and 1000 instances for two types of features: Markov features and Path features. The latter represent relations of the form "there is a directed path from $X$ to $Y$" in the PDAG of the network structure. Directed paths in the PDAG are very meaningful: if we assume no hidden variables, they correspond to a situation where $X$ causes $Y$. As discussed in Section 3, we cannot provide a closed form expression for the posterior of such a feature given an ordering. However, we can sample networks from the ordering, and estimate the feature relative to those. In our case, (we sampled 10 networks from each order). We also compared to the tradeoff curve of the *non-parametric Bootstrap* approach of [5], a non-Bayesian simulation approach to estimate "confidence" in features.

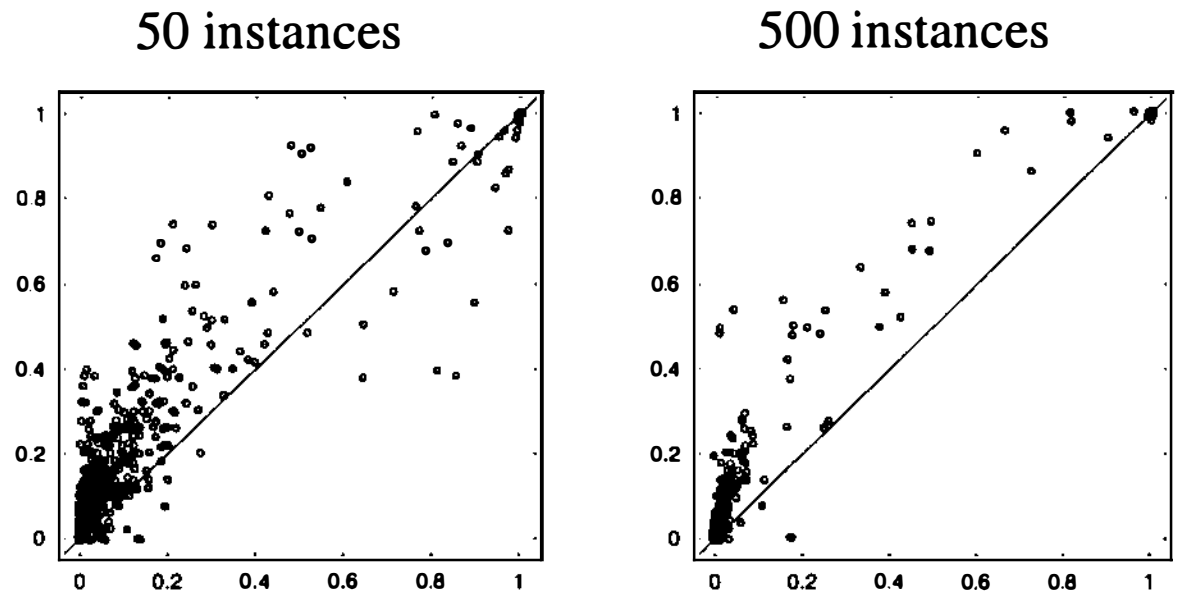

Figure 6: Scatter plots that compare posterior probability of Markov features on the *Alarm* domain as determined the two different MCMC samplers. Each point corresponds to a single Markov feature; its $x$ and $y$ coordinates denote the posterior estimated by the greedy seeded run of ordering-MCMC and structure-MCMC, respectively.

Figure 7 displays these tradeoff curves. As we can see, ordering-MCMC dominates in most of these cases except for one (Path features with 100 instances). In particular, for $t$ larger than 0.4, ordering-MCMC makes no false positive errors for Markov features on the 1000-instance data set. We believe that features it misses are due to weak interactions in the network that cannot be reliably learned from such a small data set.

## 6 Discussion and future work

In this section, we presented a new approach for estimating the true Bayesian posterior probability of certain structural network features. Our approach is based on the use of MCMC sampling, but over orderings of network variables rather than directly over network structures. Given an ordering sampled from the Markov chain, we can compute the probability of edge and Markov-blanket structural features using an elegant closed form solution. For other features, we can easily sample networks from the ordering, and estimate the probability of that feature from those samples. We have shown that the resulting Markov chain mixes substantially faster than MCMC over structures, and therefore gives robust high-quality estimates in the probability of these features. By contrast, the results of standard MCMC over structures are often unreliable, as they are highly dependent on the region of the space to which the Markov chain process happens to gravitate.

We believe that this approach can be extended to deal with data sets where some of the data is missing, by extending the MCMC over orderings with MCMC over missing values, allowing us to average over both. If successful, we can use this combined MCMC algorithm for doing full Bayesian model averaging for prediction tasks as well. Finally, we plan to apply this algorithm in our biology domain, in order to try and understand the underlying

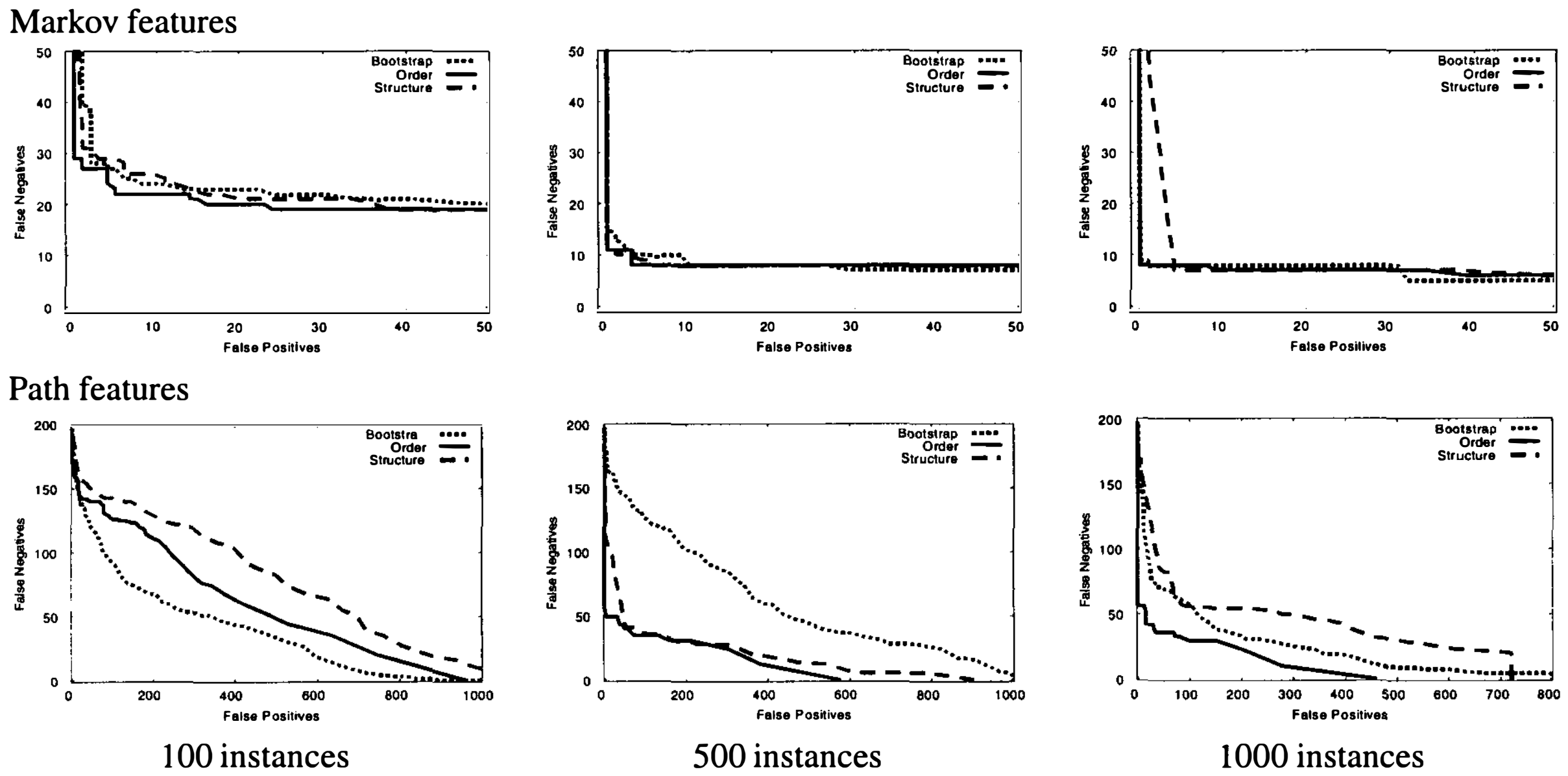

Figure 7: Classification tradeoff curves for different methods. The $x$-axis and the $y$-axis denote *false positive* and *false negative* errors, respectively. The curve is achieved by different threshold values in the range $[0, 1]$. Each curve corresponds to the prediction based on MCMC simulation with 50 samples collected every 200 and 1000 iterations in order and structure MCMC, respectively.

structure of gene expression.

**Acknowledgments** The authors thank Yoram Singer for useful discussions. This work was supported by ARO grant DAAH04-96-1-0341 under the MURI program "Integrated Approach to Intelligent Systems", and by DARPA's *Information Assurance* program under subcontract to SRI International. Nir Friedman was supported through the generosity of the Michael Sacher Trust and Sherman Senior Lectureship. The experiments reported here were performed on computers funded by an ISF basic equipment grant.